# A Series of Unfortunate Counterfactual Events: the Role of Time in Counterfactual Explanations

**Andrea Ferrario**[*]

ETH Zurich
aferrario@ethz.ch

**Michele Loi**

University of Zurich
michele.loi@uzh.ch

## Abstract

Counterfactual explanations are a prominent example of post-hoc interpretability methods in the explainable Artificial Intelligence research domain. They emerge from other types of explanations as they provide individuals with alternative scenarios and a set of recommendations to achieve a sought-after machine learning model outcome. Recently, the literature has identified desiderata of counterfactual explanations, such as feasibility, actionability and sparsity that should support their applicability in real-world contexts. However, in this paper, we show that the literature has neglected the problem of the time dependency of counterfactual explanations. We argue that, due to their time dependency and because of the provision of recommendations, even feasible, actionable and sparse counterfactual explanations may not be appropriate in real-world applications. This is due to the possible emergence of what we call "unfortunate counterfactual events." These events may occur due to the retraining of machine learning models whose outcomes have to be explained via counterfactual explanations. Series of unfortunate counterfactual events frustrate the efforts of those individuals who successfully implemented the recommendations of counterfactual explanations. This negatively affects people's trust in the ability of institutions to provide machine learning-supported decisions consistently. Finally, we introduce an approach to address the problem of the emergence of unfortunate counterfactual events that makes use of histories of counterfactual explanations. In the final part of the paper we turn to ethics and we explore two strategies organizations may adopt to cope with the challenge of unfortunate counterfactual events. We propose an ethical analysis of the two distinct strategies (and of a combination of them), focusing on a credit lending case study. We show that both strategies respond to an ethically responsible imperative to preserve the trustworthiness of credit lending organizations, the decision models they employ, and the social-economic function of credit lending.

## Introduction

The provision of explanations of machine learning model outcomes—also called post-hoc explanations—is key in the domain of explainable Artificial Intelligence (xAI) (Doshi-Velez & Kim, 2017; Lipton, 2018; Miller, 2017). Post-hoc explanations are interfaces between humans and the machine learning model that are "both an accurate proxy of the decision maker [i.e., the model] and comprehensible to humans" (Guidotti et al., 2018). They are invoked in relation to the need to 1) audit and improve machine learning models by supporting their interpretability, 2) enable learning from data by discovering previously unknown patterns, and 3) establish compliance with legislations and legal requirements (Mittelstadt et al., 2019; Samek & Müller, 2019; Watson & Floridi, 2020). Counterfactual explanations (Wachter et al., 2018) are a class of post-hoc interpretability explanations that provide the person subjected to a machine learning model-generated decision with 1) understandable information on the model outcome, and 2) a strategy to achieve an alternative (future) one. They are an example of "contrastive explanations in xAI" (Martens & Provost, 2014; Mittelstadt et al., 2019): they explain a given model outcome by sharing a "what-if" alternative scenario comprising of "feature-perturbed versions of the same persons [i.e., of the original instance]" (Mothilal et al., 2020). Recent literature from the xAI domain has discussed selected desiderata that may support the applicability of counterfactual explanations in real-world machine learning model pipelines (Laugel et al., 2017; Loi et al., 2019; Mahajan et al., 2019; Mothilal et al., 2020; Poyiadzi et al., 2020). The desiderata of feasibility, actionability and sparsity would allow to generate and share cognitively accessible counterfactual explanations that respect causal models between features, and suggest actionable strategies whose alternative scenarios comprise a limited number of features.

---
[*] corresponding author

However, at the basis of any discussion on post-hoc explanations lies the assumption that the machine learning model whose outcomes have to be explained remains "stable" or does not change, in a given time frame of interest (Barocas et al., 2020; Kroll et al., 2017; Mothilal et al., 2020).

In this paper, we argue that this assumption is violated in most real-world applications, where machine learning models are retrained with frequencies that depend on the application under consideration. We show that this phenomenon affects the provision and management of counterfactual explanations to a great extent. In fact, the time 1) of the provision of the explanation, and 2) of the successful implementation of its scenario by an interested individual are, in general, different. This time delay may lead to the emergence of unfavorable cases—called "unfortunate counterfactual events" (UCE) in these notes—where the retraining of the machine learning model invalidates the efforts of an individual who successfully implemented the scenario originally recommended by a feasible, actionable and possibly sparse counterfactual explanation.

Moreover, we introduce an approach to address the problem of the emergence of UCEs that makes use of all those counterfactual explanations that have been shared with the affected individuals until the point of time of model retraining. Finally, we discuss the emergence of UCEs from the perspective of the relation between the institution providing machine learning-generated decisions (e.g., a bank with an algorithmic credit lending system) and their post-hoc explanations, and people affected by those decisions (e.g., those people asking for a loan). We argue that series of UCEs are detrimental to trust in the institutions and in AI (Ferrario et al., 2019; Lemonne, 2018; Pasquale & Citron, 2014; Toreini et al., 2020; Zarsky, 2013). Therefore, we focus on the example of the credit lending function, and we propose an ethical analysis of two actionable strategies (and their combination). These allow preserving trustworthiness in the lending institutions, their machine learning models and the social-economic function of credit lending, while providing counterfactual explanations to potential credit borrowers.

# Counterfactual Explanations in Explainable Artificial Intelligence

## What are counterfactual explanations?

Counterfactual explanations (Wachter et al., 2018) are explanations of machine learning model outcomes that provide people with a scenario describing a state of the world—called "closest world" (Wachter et al., 2018)—in which an individual would have received an alternative machine learning outcome. For example, they explain to an individual why he or she did not receive a bank loan providing the "what-if" scenario: "you would have received the loan if your income was higher by $10,000" (Mothilal et al., 2020). This "what-if" scenario shows that an alternative outcome can be reached by altering the values of a subset of the features describing the instance at hand (i.e., the data point of the individual asking for an explanation of the denied loan, in the above example) (Mittelstadt et al., 2019; Wachter et al., 2018). For this reason, counterfactual explanations are an example of model-agnostic "feature-highlighting explanations" (Barocas et al., 2020). Not only they do provide a human-interpretable (Wachter et al., 2018) explanation of a machine learning outcome, but they outline a strategy[1] or "recommendation," to achieve an alternative, and possibly favorable, one, through the provision of a "what-if" scenario. This aspect differentiates counterfactual explanations from more descriptive machine learning model outcome explanation methods, such as Local Interpretable Model-agnostic Explanations (Ribeiro et al., 2016) and Shapley values (Štrumbelj & Kononenko, 2013). The counterfactual scenario provided by a counterfactual explanation is a "hypothetical point that is classified differently from the point currently in question" (Barocas et al., 2020). We call it "counterfactual" (data point) for simplicity. The counterfactuals are algorithmically generated (Wachter et al., 2018) by "identifying the features that, if minimally changed, would alter the output [i.e., the current outcome] of the model" (Barocas et al., 2020). In fact, the original algorithm to generate counterfactuals by Wachter et al. (Wachter et al., 2018) solves the minimization problem

$$c = argmin_c \, L(h(c), y) + d(x, c), (1)$$

where $y$ denotes the desired outcome for the counterfactual data point (i.e., alternative to the one of the original instance $x$ to be explained), $h$ is the machine learning model, $L(\cdot, \cdot)$ a loss function, and $d$ is a distance measure. The first term in (1) encodes the counterfactual condition, i.e., the search for the alternative outcome $y$, while the second term keeps the counterfactual "close" to the original instance $x$.

---

[1] In these notes, we refer to this strategy as "recommendation." The strategy is provided in the counterfactual data point, which encodes a counterfactual "scenario."

By definition, counterfactual explanations are model agnostic, providing, at the same time, a degree of protection to companies' intellectual property by the disclosure of a select set of features to third parties (Barocas et al., 2020). Moreover, they comply with legal requirements on explanations in both Europe and the United States (Barocas et al., 2020). For these reasons, they have begun to attract the interest of different sectors of society, such as businesses, regulators, and legal scholars (Barocas et al., 2020).

**Selected *desiderata* of counterfactual explanations**

Recently, the literature on counterfactual explanations has focused on selected *desiderata*, namely feasibility, actionability and sparsity (Laugel et al., 2017; Loi et al., 2019; Mahajan et al., 2019; Mothilal et al., 2020; Poyiadzi et al., 2020). These *desiderata* are deemed relevant to facilitate the applicability of counterfactual explanations in real-world applications that make use of machine learning-generated predictions. In fact, feasible, actionable and sparse counterfactual explanations recommend causality-consistent scenarios that can be implemented by the impacted individual, once he or she acts on the values of a limited number of features. Let us discuss this point in some detail.

Counterfactual explanations are said to be feasible (Poyiadzi et al., 2020; Wachter et al., 2018), if they propose a scenario that respects the causal model (Mahajan et al., 2019; Pearl, 2009) of the variables of the dataset at hand. For example, a feasible counterfactual explanation from the German Credit dataset may suggest that a loan would have been granted to an individual, if his or her income had been +10,000$, other things equal (i.e., all other variables unchanged).

On the other hand, a counterfactual scenario suggesting to decrease age, or increase the educational level from high school diploma to a master's degree without increasing age, violates causal constraints among variables. In the first case, it simply suggests an impossible recommendation. In the second case, the recommendation is not compatible with the need to spend years to achieve a Master's degree, starting from a high school diploma. We note that the Wachter et al.'s original algorithm (see equation (1)) to compute counterfactual explanations does not implement feasibility constraints. Recent approaches aim at ensuring the feasibility of counterfactual explanations by implementing post-hoc constraints on a set of generated counterfactuals. These constraints are originally introduced by domain experts to encode known causal relations between features (Mothilal et al., 2020; Poyiadzi et al., 2020).

Let us consider a feasible counterfactual explanation. We say that it is actionable (Mothilal et al., 2020; Poyiadzi et al., 2020), if the corresponding scenario can be reasonably implemented by the individual whose outcome is explained by the provision of the counterfactual explanation. Clearly, actionability is context-dependent: in particular, it depends on the capabilities of the individual implementing the counterfactual scenario. Considering Mothilal et al.'s example again (Mothilal et al., 2020), to increase the yearly income of $10,000 may be a relatively easy task for affluent individuals. However, it may represent a daunting challenge for low-income ones. Considering actionable counterfactual explanations allows excluding explanations that, although feasible, propose scenarios whose implementation is practically not realizable.[2]

Lastly, sparsity is the property of those counterfactual explanations whose scenarios suggest to alter only the values of a few variables (Laugel et al., 2017; Mothilal et al., 2020). Mothilal et al. argue that "intuitively, a counterfactual example will be more feasible if it makes changes to fewer number of features" (Mothilal et al., 2020). In other words, sparse counterfactuals "differ from the original datapoint in a small number of factors, making the change easier to comprehend" (Russell, 2019). Therefore, they are deemed to be cognitively accessible.

Sparsity becomes an important *desideratum* of counterfactual explanations, especially in big data contexts. Wachter et al. aim at ensuring sparsity of counterfactual explanations by using an L1-distance measure $d$ in (1) (Wachter et al., 2018). More recent studies have discussed sparsity by means of a two steps approach making use of a "growing spheres" algorithm (Laugel et al., 2017) and the use of a "post-hoc operation to restore the value of continuous features back to their values in x [the input data point] greedily until the predicted class [...] changes" (Mothilal et al., 2020). By definition, counterfactual explanations and the different algorithms generating counterfactuals make use of a given (and fixed) machine learning model $h$. However, in most real-world applications, machine learning models, their predictions and explanations depend on time. This basic remark has relevant repercussions on the applicability of counterfactual explanations in real-world use cases and their use as a reliable post-hoc interpretability method. We discuss this point in the forthcoming sections.

---

[2] Considering the loan example discussed by Mothilal et al., a feasible not actionable counterfactual explanation would suggest increasing the age of 20 years to an aged individual (Mothilal et al., 2020).

# The Role of Time in Machine Learning Applications

## Time and machine learning models

Machine learning models (Mitchell, 1997) are inherently dynamic objects. They are designed to perform a task, such as the binary classification of a bank's customer in "creditworthy" and "not creditworthy", by learning on data. This process is referred to as "training", or "learning" (Mitchell, 1997). After training, and depending on the application, (trained) machine learning models are deployed in IT architectures where they are fed upon batches of new data to generate predictions[3] (also referred to as outcomes) and support human decision-making. Typically, the training of machine learning models does not occur only once, i.e., just before their deployment. In fact, the process can be periodically repeated, whenever new batches of data are made available, and the performance of the machine learning model degrades. This happens as the model often generates prediction in changing environments, whose evolution is not encoded in the dataset originally used for its training. For example, in e-commerce new products become available and can be recommended on an online marketplace platform. As a result, time affects machine learning models, their predictions and the subsequent post-hoc explanations. Kroll et al. warn against a plain search for machine learning model transparency, as "systems that change over time cannot be fully understood through transparency alone" (Kroll et al., 2017). Moreover, in these cases "transparency alone does little to explain either why any particular decision was made or how fairly the system operates across bases of users or classes of queries" (Kroll et al., 2017). Considering systems with high frequency of retraining, such as algorithms serving e-commerce or social media platforms "there is the added risk that the rule disclosed is obsolete by the time it can be analyzed" (Kroll et al., 2017).

## Time to generate counterfactual explanations, and time to act on their recommendations: the case of "unfortunate counterfactual events"

As previously commented, feasible and actionable counterfactual explanations not only describe a scenario in which the user could have achieved an alternative (and preferred) outcome in understandable terms, but they highlight an actionable strategy to achieve it. In the case of sparse counterfactual explanations, this strategy focuses on altering the values of a limited number of variables. Clearly, the points of time at which 1) the explanation is generated and shared with the impacted individual,[4] and 2) its recommended scenario is successfully achieved by him or her, may differ. We argue that this time dependency of counterfactual explanations may represent a sensitive risk to their applicability.

Let us elaborate this point by describing the occurrence of an "unfortunate counterfactual event" (UCE). We refer to the Appendix for a discussion of all possible cases emerging from the analysis of the time dependency of counterfactual explanations.

In an UCE, a feasible, actionable and possibly sparse counterfactual scenario is shared with an individual who received a machine learning outcome $y_0$ at time $t_0$. The individual successfully implements the recommendations of the counterfactual scenario at a given time $t_1 > t_0$, through the investment of a certain amount of resources (*in primis*, time). Finally, at time $t_2 \geq t_1$, the machine learning model is again requested to compute a prediction for the same individual: the predicted outcome $y_1$ is the same one from $t_0$, i.e., $y_1 = y_0$.[5] In fact, in the time interval between $t_0$ and $t_1$, the machine learning model has been retrained. In particular, the retrained model did not properly learn the counterfactual scenario suggested at $t_0$, providing a wrong outcome for it. In summary, in the case of an UCE, all the efforts spent by the individual to implement the recommendations of the counterfactual explanation are frustrated by the implementation of a retrained machine learning model that did not properly learn the counterfactual and its alternative outcome.

At the time of writing, no actionable solution to the emergence of unfortunate counterfactual events has been proposed in the literature. In fact, the effects of time dependency of counterfactual explanations on their generation and provision has not yet been structurally investigated. Barocas et al. (Barocas et al., 2020) discuss four key assumptions of feature highlighting explanations, such as counterfactual explanations (Barocas et al., 2020). One assumption is that "the model is stable over time, monotonic, and limited to binary outcomes" (Barocas et al., 2020). Stability over time is not further specified, and may be interpreted as the absence of retraining or change of selected model properties. Similarly, Mothilal et al. (Mothilal et al., 2020) argue that counterfactual explanations provide the information on "what to do to obtain a better outcome in the future," (Mothilal et al., 2020), but only "assuming that the algorithm remains relatively static" (Mothilal et al., 2020). However, it is not clear how the structural stability of a model relates to the counterfactual explanations of model outcomes, and allows avoiding the occurrence of UCEs. Similarly, Verma et al. mention the dynamics of machine learning systems as a challenge to be tackled by future research on counterfactuals (Verma et al., 2020).

---

[3] In these notes, the predictions of machine learning models are also called "outcomes." In a classification problem, such as predicting whether a bank customer is creditworthy or not, the machine learning model prediction is the label "creditworthy" or "not creditworthy". We also note that most machine learning models endow their predictions with a confidence score, which is interpreted as an empirical probability.

[4] Without loss of generality, in these notes we assume that the explanations are generated and shared with an interested individual at the same point of time.

[5] In general, we note that the digital representations of individuals may change over time: some attributes are constant (e.g., the date of birth), others are slowly changing (e.g., age, education level), others may seldom change (e.g., the nationality), while others may change with high frequency (e.g., the amount of money on a credit card account). In these notes, we consider the updated data point of the individual at time $t_1$ to be equal to the (successfully implemented) counterfactual scenario to simplify the discussion.

On the other hand, Barocas et al. (Barocas et al., 2020), when discussing counterfactual explanations of credit loan, highlight that

Wachter et al. have thus argued that the law should treat a counterfactual explanations as a promise rather than just an explanation. They argue that if a rejected applicant makes the recommended changes, the promise should be honored and the credit granted, irrespective of the changes to the model that have occurred in the meantime (Wachter et al., 2018).

More specifically, to this end Wachter et al. propose:

data controllers could contractually agree to provide the data subject with the preferred outcome if the terms of a given counterfactual were met within a specified period of time (Wachter et al., 2018).

These contractual agreements are the legal counterpart of what we call the counterfactual "commitment." However, neither the nature of these contractual agreements nor their applicability (for example, considering the scenario where hundreds of customers of a company require counterfactual explanations every semester) are further discussed.

Finally, we note that Pawelczyk et al. (Pawelczyk et al., 2020) approached the problem of generating counterfactuals under predictive multiplicity (Marx et al., 2020), or the phenomenon of having multiple machine learning models with similar performance. They computed the expected cost of counterfactuals under predictive multiplicity (i.e., the minimal perturbation that would alter the label of a given data point), generalizing previous results by Ustun et al. (Ustun et al., 2019). The case of predictive multiplicity can be applied in the scenario where an existing model can be replaced by a competing one. We refer to (Pawelczyk et al., 2020) for all details.

## Honoring the counterfactual recommendation: a simple proposal

How to honor the contractual agreement with data subjects represented by the provision of a counterfactual explanation and its recommendations? As long as counterfactual explanations are implemented in a computer system at a company's premises, refraining from the retraining of machine learning models to avoid the possibility of a series of unfortunate counterfactual events is not practically feasible. As an alternative approach, we suggest to make use of the counterfactual explanations themselves, during retraining.[6]

We argue that one could perform retraining of the machine learning model on data that include, in particular, all the counterfactual scenarios that have been generated and shared with third parties until that moment, together with their alternative outcomes. The idea is that, during retraining, the model would try to learn the alternative outcomes for all the counterfactual points, together with other data deemed relevant for retraining. We show this proposal in Figure 1. The machine learning model—originally trained on data $D_{t_0}$ at $t_0$— is retrained at $t_1$, using the updated training dataset $D_{t_1}$, new data $D_{new}$, and the counterfactual scenarios $c_1, \ldots, c_n$ generated and shared with third parties until $t_1$.

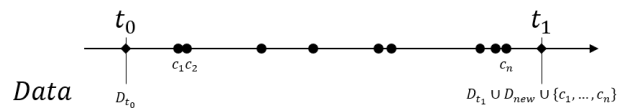

Figure 1. Counterfactual scenarios are added to the available pool of data for retraining of the machine learning model at $t_1$.

As machine learning is inherently probabilistic (Mitchell, 1997), the promise to honor the contractual agreements mentioned by Wachter et al. and represented by the counterfactual scenarios would hold only with a certain degree of certainty.[7] The quality of the agreements would be then assessed by a validation of the retrained model. Therefore, a "good" retrained model would be characterized not only by a satisfactory level of performance, but also by a high degree of certainty in honoring the counterfactual scenario agreements held up to the time of retraining. Our proposal is a form of data augmentation, which is a commonly used technique in machine learning (Antoniou et al., 2018; Shorten & Khoshgoftaar, 2019; Wei & Zou, 2019). Moreover, it is of simple implementation and does not alter the retraining algorithmic procedures by adding constraints to the optimization routines used for learning. On the other hand, by augmenting data we make use of "artificial" data points, that is the counterfactual scenarios, which may correspond to fictitious individuals (e.g., customers, patients, convicts etc.), by definition.[8] This has the effect of altering the underlying distribution and the composition of the data, i.e., a portfolio of customers, cohorts of patients or convicts, used to retrain the model. However, we argue that, although fictitious, the data points encoding the counterfactual scenarios, may represent *possible* individuals. These become *certain* to the institution using the machine learning model, at the moment at which the individual impacted by the original counterfactual explanation successfully implements the recommended scenario and requests a new prediction for himself or herself to the model.

---

[6] We do consider only scenarios of feasible and actionable counterfactual explanations, in what follows.

[7] As noted in footnote 3, each machine learning outcome is endowed with a score $p$ (normalized in the interval [0,1]) that is interpreted as a probability, or degree of certainty in the corresponding outcome. Therefore, in the case of a counterfactual scenario, the corresponding outcome is honored with degree of certainty $p$.

In view of the above data augmentation approach, we argue that the ontological status of counterfactual scenarios of feasible and actionable counterfactual explanations advocates for their use in the retraining of machine learning models. Therefore, the correct estimation of their outcomes by a retrained machine learning model becomes the degree of protection from the aforementioned series of unfortunate counterfactual events.

Depending on the application, the class distribution of training data, e.g., the number of creditworthy vs. not creditworthy customers in the credit lending case, may be imbalanced.[9] As an effect, considering counterfactual scenarios and their outcomes in the training data at time $t_1$ may increase class imbalance.[10] In that case, standard machine learning techniques to cope with class imbalance, such as class-weighted learning, sub-sampling or over-sampling (Fernández et al., 2018; He & Ma, 2013; Krawczyk, 2016) may be taken into account.

Finally, we argue that the probability of a counterfactual scenario to be successfully implemented by an individual correlates positively with the actionability of the corresponding counterfactual explanation. Therefore, companies could envisage the development of scoring systems to assess the probability of successful implementation of counterfactual scenarios, once a sufficient number of cases are collected. This information could be used to assist the validation of the retrained machine learning models, in order to select those models that correctly learn counterfactual scenarios with high implementation probabilities.

Finally, we mention that the use of counterfactuals to augment datasets has been recently proposed by Kaushik et al. (Kaushik, Hovy, et al., 2020; Kaushik, Setlur, et al., 2020), although in the context of natural language processing (e.g., sentiment analysis of movie reviews on online platforms). To this end, Kaushik et al. proposed to tackle the problem of model reliance on spurious correlations by letting human editors to edit sampled documents (e.g., movie reviews) "to render (designated) counterfactual labels applicable" (Kaushik, Setlur, et al., 2020).

Their results showed that the models trained on the augmented datasets that included counterfactuals are less sensitive to spurious correlations and with a high out-of-sample performance on different datasets (Kaushik, Hovy, et al., 2020; Kaushik, Setlur, et al., 2020).

## Ethical Implications and Trade-Offs in Committing to Counterfactual Explanations

The availability of models that are consistent with sets of counterfactual decisions is of the utmost importance each time exogenous factors impact models performance to the extent that their retraining becomes necessary. A case in point is the recent COVID 19 pandemic that entailed abrupt changes in economic expectations. These may lead to an overall disruption of financial operations such as credit lending.

One may imagine different options to preserve the trustworthiness of an institution facing these operation-impacting changes and, at the same time, committing to counterfactual scenarios.[11] In the forthcoming sections we introduce two

---

[8] In general, these artificial data points may, or may not, belong to the dataset used to retrain the machine learning model, in the first instance. Therefore the adjective "artificial." However, this depends on the specific implementation of the algorithmic search for counterfactuals as in equation (1). For example, all counterfactuals would belong to the training dataset (therefore, being "artificial" no more), if the search for counterfactuals is limited to the samples in the training dataset with an outcome alternative to the one of the original instance to be explained. This is the case of Algorithm 1 in (Laugel et al., 2017). No such constraint is mentioned in Wachter et al.'s paper (Wachter et al., 2018).

[9] Typically, in credit lending applications, the number of not creditworthy customers is smaller than the number of creditworthy ones. For example, in the German Credit dataset, not creditworthy customers represent 30% of all data.

[10] If the distribution of classes in training data from time $t_0$ to $t_1$ does not vary significantly. Otherwise, it is not possible, a priori, to infer how the class imbalance in data may be affected by taking into consideration counterfactual scenarios and their outcomes. This would be the case in an economic downturn scenario, such as the one discussed in Section 4.

[11] The violation of counterfactual commitments is a source of reputational risk for the institution. For example, in the case of a bank it may lead to customer churn, with an effect on the profitability of the portfolio that depends, among others, on the number and the reserves of the customers impacted by the violation of the commitments.

main options that involve distinct ethical trade-offs and we comment on their applicability.

In what follows, we assume that trustworthiness is compromised by a violation of promise keeping, and that trust is compromised when the entrusting agent is aware of the violation. This is compatible with a strand in the literature that considers the *goodwill* of the trusted a condition for its trustworthiness and the assumption of this goodwill by the trustor as a cognitive aspect of trust (Baier, 1991). For violating a promise is arguably incompatible with the goodwill of the trusted in the type of scenarios we consider.

**Option I: declaring boundary conditions for counterfactual commitments**

The first approach for the institution is to keep only counterfactual commitments within well-specified "boundary conditions," i.e., the circumstances within which the explanation can be treated formally as a promise. These circumstances are applied to all counterfactual commitments in a given time frame,[12] and may appeal to understandable values and constraints such as economic necessity.

We stress out that the greatest threat to trustworthiness here is having undisclosed boundary conditions motivating the decision-maker to break its (implicit or explicit) counterfactual commitment. This can be avoided if the organization is transparent with the recipient of the explanation about the boundary conditions in order. For example, the organization may identify and publicize a given set of circumstantial conditions under which it will make decisions that depart significantly from the previously communicated counterfactuals. For example, a bank may have a criterion stating explicitly that it will not be able to stick to its counterfactual commitments in the event of a downward or upward movement of a certain economic index higher than a given threshold, or when honoring the commitment would imply a profit loss higher than a given threshold. To be trustworthy, the company would need to refer to either public data (e.g., a publicly accessible economic index) or confidential data that can be audited confidentially by trustworthy independent parties (the auditors may need to obtain access to confidential information to be able to verify if certain claims, e.g., about prospects of economic losses, are credible). When the decision-maker is confident about its ability to avoid exceptional circumstances (or exceptional choices in exceptional circumstances) within a given period, the former strategy can be realized by simply stating counterfactual commitments with an "expiration date."

We would like to point out that adopting circumstance-limited or time-limited counterfactual commitments is ethically valuable, i.e., valuable impersonally, for society at large, and it is even a moral duty. We argue that this is not only in the self-interest of the institution. To see why, let us consider the case of a "sudden economic downturn scenario" for a bank, or, in general, a credit lending institution.

In this scenario, fulfilling all extant counterfactual commitments likely implies awarding loans that are too risky. One immediate consequence of a defaulting client is a cost for the lender, as defaulted loans reduce overall profit. However, due to the economic downturn, a significant proportion of defaulting clients may well cause the financial collapse of the institution. Such behavior is clearly unethical on Kantian deontological grounds: if all credit institutions acted according to this moral maxim (i.e., lending to all clients fulfilling past counterfactual explanations, even when the updated model predicts them to default), the likely result would be the collapse of the entire financial system, so no credit would be possible. This is a violation of Kant's categorical imperative (Kant, 2002). Interestingly, violating counterfactual commitments also violates a common-sense deontological rule against breaking promises (which also violates Kant's categorical imperative).

In summary, stating clearly that counterfactual commitments are only valid within situational constraints becomes a viable alternative to the following ethically permissible, although limiting, options: a) avoiding the provision of counterfactual explanations altogether, or b) avoiding the interpretation of counterfactual explanations as implicit promises on the recipient's end.

**Option II: a probabilistic approach to counterfactual commitments**

As previously pointed out, the proposal of considering past counterfactual scenarios during the retraining of machine learning models will typically not lead to the fulfillment of all past counterfactual commitments. In fact, as a result of the use of counterfactual scenarios during retraining, the corresponding counterfactual commitments are endowed with degrees of certainty (see Section 3). This said, in the probabilistic approach an institution may decide to guarantee that a subset of counterfactual explanations and subsequent commitments made at time $t_0$ will hold at time $t_1 > t_0$, based on statistical reasoning.

For example, a bank could guarantee only those commitments with a degree of certainty greater than a given threshold (this is equivalent to state to third parties that each commitment has a degree of risk, which is quantified using the threshold). As noted in Section 3, the degree of certainty of counterfactual scenarios is computed as result of the machine learning model retraining, i.e., only after the generation of the corresponding counterfactual explanation (at time $t_0$). Therefore, it cannot be considered as a clause of validity for the commitment at $t_0$. On the other hand, the same bank may decide to guarantee only a fixed percentage of a given type of commitment (e.g., those involving the increase of annual income), based on the statistical analysis of histories of past ones.

---

[12] Therefore, the choice of the term "boundary conditions."

However, in general, institutions will not be able to tell, for a specific client, whether his or her specific counterfactual commitment will be maintained. Even if a company relies on a public (or independently auditable), systematic, non-morally arbitrary way to select the claims to be fulfilled in altered circumstances, there may be no way to know in advance which claims fulfill those conditions. For different commitments may be unequally hard to satisfy in different circumstances, and, at the time in which counterfactual explanations are given, it will not be known in advance what are the low probability circumstances that will occur and call for a revision of the promises made. In other words, at $t_0$ one may argue that A's counterfactual commitment could be easier to satisfy than B's. However, at $t_1 > t_0$ the reverse may be true.[13]

If the probabilistic approach has to be trustworthy, it must rely on a public (or independently auditable), systematic, non-morally arbitrary way of selecting the counterfactual commitments to be discarded vs. those to be maintained using statistical reasoning. The advantage of this strategy is that it can help the institution providing these commitments to find the optimal balance between two conflicting prima facie moral obligations: a) to fulfill the expectation of the counterfactual commitment, and b) to enable mutually beneficial and socially advantageous economic transactions that require violating past commitments.

The probabilistic strategy provides a "grey option" alternative to the option of being trustworthy but risking institutional collapse (e.g., bankruptcy), and the option of violating past commitments and doing simply what one has most reason to do, given future prospects and ignoring the history of the interaction. The grey option is a quantitative constraint on the breaking of promises. It limits the damage to trustworthiness and deontological morality, by limiting commitment breaking to a given percentage of cases, which is planned, known in advance, and controlled by the decision maker.

The difference with the first strategy is that, in the first, one refrains from the commitment altogether, declaring it is not valid under specific circumstances. Here, instead, commitments are made and then (selectively, and partially) broken. The expectations of some clients are upset, because these clients could not know that their counterfactuals would no longer be valid. However, the damage is limited, because only a limited proportion of expectations is violated.

Framed in this way, the solution ignores deontological morality altogether. The goal is to obtain the desired balance of trustworthiness and profitability (and sustainability) and, in order to achieve the desired mix, the company decides to break promises, but only a certain degree.

This may seem hypocritical, but it is morally justifiable from a consequentialist standpoint. A credit institution, for example, faces a reputational dilemma between a) disrespecting its commitments to clients implied by counterfactual explanations, with negative impact on trust in the system, and b) eroding profitability or worse engendering the financial collapse of the institution. By training models that are probabilistically constrained by past counterfactuals, and using this information to determine statistical rules to guarantee counterfactual commitments, this dilemma can be turned into a trade-off the institution can control. A bank, for example, may find it optimal to sacrifice some degree of profitability for the sake of promoting trust from the majority of clients. Faced with a choice between honoring past commitments and financial stability in the long run, the probabilistic strategy achieves the latter at the expense of the former. The idea of reducing the amount of broken promises, rather than fulfilling all promises, also has a distinctive flavour of "negative consequentialism." In fact, consequentialism can be stated in a negative form: to act so as to prevent the most harm or most moral wrongs (Savulescu, 1998). If the company honors all counterfactual commitments, when it can no longer afford to do so, it risks going bankrupt, which implies the failure to honor commitments to shareholders and other stakeholders that are harmed by bankruptcy. Therefore, by selectively discarding some, the bank minimizes the proportion of commitments it breaks in the long term.

While the morality of this approach may be doubted by strict deontologists, utilitarianism recommends it. The option is morally objectionable because it sacrifices moral obligations to a minority of clients for the sake of higher-order long-term objectives of trustworthiness combined with financial sustainability. But if you are utilitarian, this is a bullet you are willing to bite anyway, and for the general case.

---

[13] E.g., at $t_0$ John is unlikely to default on debt because he runs a highly profitable pub since ten years. However, at $t_1$ and during the Covid 19 pandemic, John is no longer able to generate a stable income.

**Option III: combining declared boundary conditions and probabilistic approaches**

Let us suppose that the objection from deontological morality against the probabilistic approach should be taken seriously. Can the probabilistic approach be made compatible with deontological morality? The answer is affirmative and it consists in combining the declared boundary condition and probabilistic approaches. The core of the combined approach is that the probabilistic nature of the commitment to respect counterfactuals should be honestly and transparently declared to the clients. This is in the interest of both bank and clients: from the point of view of the bank, it prevents the machine learning model to behave fully erratically. The clients, on the other hand, can rationally factor the risk of an unfulfilled counterfactual commitment in their action plans. When making a financial decision, the client will not be forced to decide "should I trust the counterfactual explanation or not?", but can reasonably ask "how much should I trust the counterfactual explanation?", and decide that also based on the probability of realization attached to it.

Let us consider the economic downturn scenario again. In the case of a sudden downturn, a bank may be able to approve lending to a known proportion of individuals who fulfilled the counterfactual condition given in the past (but that no longer satisfy the relevant risk metrics after the model update), where fulfilling this condition for all would have meant going bankrupt. The financial risk arising from fulfilling a known proportion of past counterfactuals can be approximately assessed in advance, given that the selection of counterfactuals to be disrespected is grounded in a statistical model. Ideally, a bank will be able to cover the cost arising from the need to promote its trustworthiness in a sustainable manner (e.g., by raising the interest rate on safer loans, or distributing the costs among many different stakeholders).

The bank in the example can also plan in advance to disrespect a maximum proportion of counterfactual commitments in case a specified type of future risk will materialize itself. This plan violates deontological morality when it is kept secret and allows clients to form misplaced expectations in the future behavior of the bank. But it could also not violate it if it is communicated at the time the explanation is given. Moreover, keeping the plan secret ignores the potential benefit for the client of accessing the probabilistic information the bank has. Knowing the probability that a promise will not be respected is valuable in the client's perspective. Ideally, a rational client would want the client to be fully aware of the probabilistic uncertainty of a probabilistic commitment to a counterfactual, in order to know who much weight to place on it when making decisions.
Hence, one may solve the tension between the consequentialist and the deontological approaches by declaring the probabilistic nature of the plan to respect counterfactual, to the client, in the moment the explanation is given to him or her. It may be objected that "making a promise while declaring that it may be broken" would also violate Kant's categorical imperative, and count as unethical in that perspective. This is true of promises that are made with the intention that they may be broken for arbitrary reasons. This is not what the combined strategy does. To probabilistically guarantee a proportion of counterfactuals, in a way that varies according to the circumstances, based on a plan established in advance, and that is honestly communicated to the recipient of the explanation, amounts to a commitment that has social value. Finally, let us return to our credit bank scenario to see how that may play out in reality. A bank may provide a client with the following prospect: "we expect ourselves to behave as described in our explanation in 90% of cases except in economic circumstances X or Y where that proportion falls to 70% and we offer no guarantee about honoring our commitments when Z occurs." This would be a promise that includes the conditions of its own breaking, with a probability attached to it. Like ordinary promises, this probabilistic and conditional commitment can ground reasonably accurate expectations of the client about the bank's behavior. The recipient of the counterfactual commitment receives useful information, and knows what to expect in ordinary circumstances, factoring it some degree of risk. We could label this approach as wisdom-by-design: the system will exhibit a certain degree of coherence and predictability, yet avoid rigidity in its commitment to the past.

## Conclusions

Counterfactual explanations are a class of contrastive explanations of machine learning outcomes with interesting properties. In fact, the possibility of suggesting a strategy to have recourse against a machine learning model outcome is a useful tool available to those affected by AI-assisted decisions. However, we showed that this property becomes a double-edged sword in the hands of those organizations that make use of AI in their decision-making processes, due to the role of time in machine learning applications. The approach to tackle the emergence of series of unfortunate counterfactual events consists in 1) using counterfactual scenarios to augment the dataset used to train the models, and to 2) enforce ethical trade-offs by different sets of constraints on the commitments of the counterfactual scenarios. This approach is a first step towards a systematic analysis of the methods to 1) ensure the consistent use of counterfactual explanations in real-world applications, and 2) support trust in institutions, and their AIs, while striving for the interpretability of machine learning models. We advocate for qualitative and empirical studies with real datasets to test the propensity of organizations in implementing different ethics-preserving strategies to commit to counterfactual explanations, and to assess the impact of series of unfortunate counterfactual events on their functions.

# Appendix

In Table 1, we enumerate all possible cases that emerge from the change in time of data points, machine learning models and their outcomes, when considering the implementation of counterfactual scenarios. We show that the unfortunate counterfactual event is one of these cases (i.e., case 4). We start with some notation: with $x$ we denote a given data point corresponding, for example, to a specific individual. With $h$ we denote the machine learning model at hand (e.g., a credit loan system). The outcome of $x$ computed by $h$ is $y$, or $y = h(x)$. Let us consider two distinct moments of time $t_0$ and $t_1$, with $t_1 > t_0$. In Table 1 we write "+" if the corresponding element does not change from $t_0$ to $t_1$, "-" otherwise. In case a "-" is highlighted for the data point x, the counterfactual scenario of the explanation at $t_0$ is implemented at $t_1$. It follows that we need to describe 8 distinct cases.

We note that the occurrence of an unfortunate counterfactual event is given by case 4. On the other hand, case 6 is a "paradigmatic counterfactual event:" the implementation of the counterfactual scenario is successful and occurs in the absence of machine learning model changes. Case 8 encodes the possibility of having a machine learning model retraining that is compatible with the implementation of a counterfactual scenario (i.e., ending in the change of outcome $y$, as desired). All other cases are not relevant for the implementation of counterfactual scenarios we discuss in these notes.

| Case | $x$ | $h$ | $y$ | Description |
|---|---|---|---|---|
| 1 | + | + | + | No change. No counterfactual scenario is applied, in particular. |
| 2 | - | + | + | Only $x$ changes, although its change does not alter the corresponding prediction. This case cannot hold if the individual has successfully implemented the counterfactual scenario suggested at time $t_0$ as, in that case, the outcome at $t_1$ would have changed (given the same machine learning model $h$). |
| 3 | + | - | + | The change of the world reflected in the retraining of $h$ does not result in a change in outcome for $x$. This case is not relevant for the discussion on the implementation of counterfactual scenarios, as in that case we would have had a change in $x$. |
| 4 | - | - | + | Unfortunate counterfactual event. |
| 5 | + | - | - | This case is not relevant to discuss the implementation of counterfactual scenarios, as the data point $x$ does not change. The outcome $y$ changes due to the possible retraining of the machine learning model $h$. |
| 6 | - | + | - | Paradigmatic counterfactual event: the data point $x$ changes due to the implementation of the counterfactual scenario, resulting in the change of the outcome $y$, other things equal. |
| 7 | + | + | - | Not applicable |
| 8 | - | - | - | The change of the data point $x$ due to the implementation of a counterfactual scenario and of the machine learning model $h$ are "compatible," as the outcome $y$ changes as well. |